\documentclass{aircc}
\usepackage{mathpartir}
\usepackage{hyperref}
\usepackage{listings}
\usepackage{lmodern} 
\usepackage{diagbox}

\usepackage{amssymb}
\usepackage{graphicx}
\usepackage{url} 
\usepackage{amsmath}
\usepackage{color}
\usepackage{epsfig}
\usepackage{amsfonts}
\usepackage{latexsym}
\usepackage{mathtools}
\usepackage{multirow}
\usepackage{multicol}

\usepackage{algorithm}
\usepackage{algorithmic}
\usepackage{tikz-qtree}
\usepackage[font=small,labelfont=bf]{caption}
\usepackage[footnote]{acronym}
\usepackage{lscape}
\usepackage{url}
\usepackage{extarrows}
\usepackage[percent]{overpic}
\usepackage{wrapfig}
\usepackage{soul}
\usepackage{comment}
\usepackage{tikz}
\usetikzlibrary{positioning}
\usetikzlibrary{shapes.geometric, arrows}

\usepackage{amsmath,amsfonts,amsthm}
\usepackage{subfig}
\usepackage{bold-extra}

\newcommand{\bqs}{\begin{eqnarray*}}
\newcommand{\eqs}{\end{eqnarray*}}
\newcommand{\bq}{\begin{eqnarray}}
\newcommand{\eq}{\end{eqnarray}}
\newcommand{\biz}{\begin{itemize}}
\newcommand{\eiz}{\end{itemize}}
\newcommand{\be}{\begin{enumerate}}
\newcommand{\ee}{\end{enumerate}}

\begin{document}

\title{
An Enhanced Ad Event-Prediction Method  Based on Feature Engineering
}

\author{Saeid Soheily-Khah and Yiming Wu}
\affiliation{SKYLADS Research Team, Paris, France \\
\email{\{saeid,yiming\}@skylads.com}}

\maketitle

\begin{abstract}
In digital advertising, Click-Through Rate (CTR) and Conversion Rate (CVR) are very important metrics for evaluating ad performance.
As a result, ad event prediction systems are vital and widely used for sponsored search and display advertising as well as Real-Time Bidding (RTB).
In this work, we introduce an enhanced method for ad event prediction (i.e. clicks, conversions) by proposing a new efficient feature engineering approach. 
A large real-world event-based dataset of a running marketing campaign is used to evaluate the efficiency of the proposed prediction algorithm.
The results illustrate the benefits of the proposed ad event prediction approach, which significantly outperforms the alternative ones.
\end{abstract}


\section{Introduction}
Ad event prediction is critical to many web applications including recommender systems, web search,  sponsored search, and display advertising \cite{Cheng:2010:PCP:1718487.1718531,Zhang:2014:SCP:2893873.2894086,Chapelle:2014:SSR:2699158.2532128,Borisov:2016:NCM:2872427.2883033,Cheng:2016:WDL:2988450.2988454}, and is a hot research direction in computational advertising \cite{Li:2015:CPA:2783258.2788582,Chen:2016:DCP:2964284.2964325}. 
The event prediction is defined to estimate the ratio of events such as videos, clicks or conversions to impressions of advertisements that will be displayed.
Generally, ads are sold on a 'Pay-Per-Click' (PPC) basis or even 'Pay-Per-Acquisition' (PPA), meaning the company only pays for ad clicks, conversions or any other pre-defined actions, not ad views. 
Hence, the Click-Through Rate (CTR) and the Conversion Rate (CVR) are very important indicators to measure the effectiveness of advertising display, and to maximize the expected value, one needs to predict the likelihood that a given ad will be an event, in the accurate way possible.
As result, the ad prediction systems are essential to predict the probabilities of a user doing an action on the ad or not, and the performance of prediction model has a direct impact on the final advertiser and publisher revenues and plays a key role in the advertising systems.
However, due to the information of advertising properties, user properties, and context environment, the ad event prediction is very fancy, challenging and complicated, 
and is a massive-scale learning problem.

In the multi-billion dollar online advertising industry, mostly all web applications relied heavily on the ability of learned models to predict ad event rates accurately, quickly, and reliably \cite{Richardson:2007:PCE:1242572.1242643,citeulike:4375063,Graepel:2010:WBC:3104322.3104326}.
Even 0.1$\%$ of improvement in ad prediction accuracy would yield greater revenues in the hundreds of millions of dollars.
While, with over billions daily active users and over millions active advertisers, a typical industrial model should provide predictions on billions of events per day.
Hence, one of the main challenges lies in the large design space to address issues of scale. In the case, for ad event prediction, we need to rely on a set of well-designed features.
However, to capture the underlying data patterns, selecting and encoding the proper features has also pushed the field.

In this work, we discuss the machine learning methods for ad  event prediction and  propose a dedicated feature engineering procedure as well as a new efficient approach to predict more effectively whether an ad will be an event or not. 
The novel enhanced algorithm mainly facilitates feature selection using the proposed statistical techniques, thereby enabling us to identify, a set of relevant features.
%
The remainder of this paper is organized as follows. 
In Section 2, we present briefly the state-of-the-art of a variety of classification techniques which widely used for ad event prediction applications.
Next, we describe machine learning and data mining methods for feature engineering including feature selection, feature encoding and feature scaling. 
In Section 4,  we describe the proposed feature engineering strategy being directly applicable in any event prediction system. 
Finally, the conducted experiments and results obtained are discussed in Section 5, and Section 6  concludes the paper. The main contributions of this research work are as follows:
\begin{itemize}
    \item[-] We propose two novel adjusted statistical measures for feature selection.
    \item[-] We provide an enhanced framework  for ad event-prediction by analyzing the huge amount of historical data. The introduced framework includes the pipelines for data pre-processing, feature selection , feature encoding, feature scaling, as well as training and prediction process.
    \item[-] We show through a deep analysis of a very large real-world dataset, that the proposed strategy significantly outperform the alternative approaches.
\end{itemize}

In the remainder of the paper, specially in the experimental study, to simplify, we consider the events as clicks. Of course, all the design choices, experiments and results can however be directly extended to any other events such as conversions.

\section{State-of-the-art}
In the literature, a  variety  of  classification  techniques  such  as  logistic regression, support vector machine, (deep) neural  network, nearest neighbor, naive Bayes, decision tree and random forest  have  been  widely  used  as machine learning and data mining techniques for ad event prediction applications.

Logistic regression contains many techniques for modeling and analyzing several features, when the focus is on the relationship between a dependent feature and one (or more) independent features. 
More specifically, the regression analysis is a statistical process which helps one understand how the typical value of the dependent feature changes when any one of the independent features is varied, while the other ones are kept fixed.
In the literature, (logistic) regression model have been used by many researchers to solve the ad event prediction problems for advertising \cite{Chapelle:2014:MDF:2623330.2623634,Effendi2017ClickTR,41159,Richardson:2007:PCE:1242572.1242643}.

Gradient boosting is one of the most powerful machine learning algorithms, 
which produces a prediction model in the form of hybrid weak models, typically  decision trees \cite{Friedman:2002:SGB:635939.635941}.
The boosting notion came out of the idea of whether a weak learning model can be modified to become better.
It builds the model in a stage-wise manner like other boosting methods do, and  generalizes them by allowing the optimization of a loss function.
Gradient boosting represents 'gradient descent' plus 'boosting', where the learning procedure sequentially fits novel models to provide a more accurate response estimation. 
In simple words, the principle idea behind this method is to construct the novel base-learners to have maximal correlation with the negative gradient of the loss function, associated with the whole hybrid model.
Gradient boosting technique,  practically, is widely used in many prediction applications due to its easy use, efficiency, accuracy and feasibility \cite{Richardson:2007:PCE:1242572.1242643,Trofimov:2012:UBT:2351356.2351358}, as well as the learning applications \cite{burges2010ranknet,Dave:2010:LCR:1835449.1835671}.

Bayesian classifiers are statistical methods that predict class membership probabilities.
They works based on the Bayes' rules (alternatively Bayes' law or Bayes' theorem), where features are assumed to be  conditionally independent. 
Even in spite of assumption of the features dependencies, in practice, they provide satisfying results, they are very easy to implement and fast to evaluate, where they just need a small number of training data to estimate their parameters. 
However, the main disadvantage is where the Bayesian classifiers make a very strong assumption on the shape of data distribution. 
In addition, they can not learn interactions between the features, and  suffer from zero conditional probability problem (division by zero) \cite{Domingos1997}, where one simple solution would be to add some implicit examples. 
Furthermore, computing the probabilities  for continuous features  is not possible by the traditional method of frequency counts.
Nevertheless, some studies have found that, with an appropriate pre-processing, Bayesian classifiers can be comparable in performance with other classification algorithms \cite{Graepel:2010:WBC:3104322.3104326,journals/jcit/Entezari-MalekiRM09,journals/bmcbi/KukrejaJS12,Lorena20115268}. 

Neural networks are modeled based on the same analogy to the  human  brain  working, and  are  a  kind  of  artificial intelligence based methods for ad event prediction problems  \cite{Zhou2017DeepIN}. 
Neural networks algorithms benefit from their learning procedures to learn the relationship between  inputs  and   outputs by adjusting the network
weights and biases, where the weight refers to strength of connections between two units (i.e. nodes) at different layers.
Thus,  they  are  able  to  predict the accurate class label of the input data. 
In \cite{Liu:2015:CCP:2806416.2806603}, authors extended Convolutional Neural Network (CNN) for click prediction, however they are biased towards the interactions between neighboring features. 
Most recently, in \cite{Zhang2016DeepLO}, authors proposed a factorization machine-supported neural network algorithm, 
to investigate the potential of training neural networks to predict ad clicks based on multi-field categorical features.
However, it is limited by the capability of factorization machines. 
In general, among deep learning frameworks for predicting ad events, Feed Forward Neural Networks (FFNN) and Probabilistic Neural Network (PNN) are claimed to be the most competitive algorithms.

Lastly, random forest \cite{Breiman2001} is  an ensemble learning approach  for classification, regression and some other tasks such as estimating the feature importance,  which operates  by  constructing  a  plenty  of decision  trees.
Particularly, random forest is a combination of decision trees that all together produce predictions and deep intuitions into the data structure. 
While in standard decision trees, each node is split using the best split among all the features based on the Gini score, while in a random forest, each node is split among a small subset of randomly selected input features to grow the tree at each node. 
This strategy yields to perform very well in comparison with many other classifiers such as support vector machine, neural network and nearest neighbor.
Indeed, it makes them robust against the over-fitting problem as well as an effective tool for classification and prediction \cite{liaw02:rf_pack,8367767}. 
However, applying decision trees and random forests to display advertising, has additional challenges due to having categorical features with very large cardinality and the sparse nature of the data, in the literature, many researchers have used them in predicting ad events \cite{Dembczynski2008PredictingA,Trofimov2012,shi2016}.

Nevertheless, one of the most vital and necessary steps in any event prediction system is to mine and extract features that are highly correlated with the estimated task. Moreover, many experiment studies are conducted to show that the feature engineering improves the accuracy of ad event prediction systems.
The traditional event prediction models mainly depend on the design of features, while the features are artificially selected, encoded and processed. 
In addition, many successful solutions in both academia and industry rely on manually constructing the synthetic combinatorial features \cite{Shan:2016:DCW:2939672.2939704,soheily2018ensemble}.
Because, the data sometimes has a complex mapping relationship, and taking into account the interactions between the features is vital.
In the following, we discuss about state-of-the-art of the feature engineering approaches, which can be considered as the core problem to online advertising industry, prior to introduce our proposed approach in feature engineering and event prediction.

\section{Feature engineering}
\label{sec:featureeng}
Feature engineering is the fundamental to the application of  machine learning, data analysis and mining as well as mostly all artificial intelligence tasks, and generally, is difficult, costly and expensive.
In any artificial intelligence or machine learning algorithm (e.g. predictive and classification models), the features in the data are vital, and they dramatically influence the results we are going to achieve. Therefore, the quantity and quality of the features (and data) have huge influence on whether the model is good or not.
In the following, we discuss the data pre-processing process and feature engineering in more detail and we present the most well-used  methods in the case.

\subsection{Feature selection}
Feature selection is the process of finding a subset of useful features and removing irrelevant features to use in the model construction.
It can be used for a) simplification of models to make them easier to expound, b) reduce training time consumption, c) avoid the curse of dimensionality and etc.
In simple words, feature selection determines the accuracy of a model and helps remove useless correlation in the data that might diminish the accuracy.
In general, there are three types of feature selection algorithms: filter methods, wrapper methods and embedded methods.

Embedded approaches learn which features best contribute to model accuracy while it is being created. 
It means that some learning algorithms carry out the feature selection as part of their overall operation, such as random forest and decision tree, where we can evaluate the importance of features on a classification task.
The most common kind of embedded feature selection approaches are regularization (or penalization) methods, which inset additional restrictions into the optimization of a predictive algorithm.
However, creating such a good model is challenging due to the computational cost and time of model training.

Wrapper methods consider the selection of a subset of useful features as a search problem, where different combinations are constructed, evaluated and then compared to other ones. 
A predictive model used to assess the combination of features and assign a score based on the accuracy of the model, where the search process could be stochastic, methodical, or even heuristic.

Filter feature selection approaches apply a statistical test to assign a goodness scoring value to each feature. The features are ranked by their goodness score, and then, either selected to removed from the data or to be kept. 
These filter selection methods are generally univariate and consider the feature independently (e.g. chi-square test),
or in some cases, with regard to the dependent feature (e.g. correlation coefficient scores).

\subsection{Feature encoding}
In machine learning, when we have categorical features, we often have a major issue: how to deal with categorical features?
Practically, one can postpone the problem using a data mining or machine learning model which handle the categorical features (e.g. $k$-modes clustering), or deal with the problem (e.g. label encoding, one-hot encoding, binary encoding)

When we use such a learning model with categorical features, we mostly have three types of models: a) models handling categorical features accurately, b) models handling categorical features incorrectly, or c) models do not handling the categorical features at all.

Therefore, there is a need to deal with the following problem.
Feature encoding points out to transforming a categorical feature into one or multiple numeric features. One can use any mathematical or logical approach  to convert the categorical feature, and hence, there are many methods to encode the categorical features, such as: a) numeric encoding, which assigns an arbitrary number to each feature category, b) one-hot encoding which converts each categorical feature with $m$ possible values into $m$ binary features, with one and only one active, c) binary encoding to hash the cardinalities into binary values, d) likelihood encoding to encode the categorical features with the use of target (i.e. label) feature.  From a mathematical point of view,  it means a probability of the target, conditional on each category value, and e) feature hashing, where a one-way hash function convert data into a vector (or matrix) of features. 

\subsection{Feature scaling}
Most of the times, the data will contain features highly varying in units, scales and ranges.
Since, most of the machine learning and data mining algorithms use Eucledian distance between two data points in their computations, this makes a problem.
To suppress this effect, we need to bring all features to the same level of unit, scale or range. This can be attained by scaling.
Therefore, feature scaler is a utility that converts a list of features into a normalized format suitable for feeding in data mining and learning algorithms.
In practice, there are four common methods to perform feature scaling:
a) min-max scaling to rescale the range of features in [0, 1] or [−1, 1], b) mean normalisation to normalize the values between -1 and 1, c) standardisation, which swaps the values by their $Z$ scores, and d) unit vector, where feature scaling is done in consideration of the entire feature vector to be of unit length.

Notice that, generally, in any algorithm that computes distance or assumes normality (such as nearest-neighbor), we need to scale the features, while feature scaling is not indispensable in modeling trees, since tree based models are not distance based models and can handle varying scales and ranges of features. 
As more examples, we can speed up the gradient descent method by feature scaling, and hence, it could be favorable in training a neural network, where doing a features scaling in naive Bayes algorithms may not have much effect.

\section{The design choices}
The proposed feature engineering strategy is briefly presented in to the following steps (see Algorithm \ref{algo1}), where in the reminder of this section, we explain in detail the proposed feature learning approach for the ad event prediction.

Typically, there are plenty of recorded information, attributes and measures in an executed marketing campaign. For instance, the logs services enable advertisers to access the raw, event-level data generated through the online platform in different ways. However,  we are not interested in all of them. Lots of recorded information and data are not useful or available for us, even they increase the complexity. So, at the first step, we prune the raw data, before doing any mining task.

\begin{algorithm}[H]
 \caption{The proposed feature engineering strategy}
 \label{algo1}
\begin{algorithmic} 

\STATE \textbf{input}:  {$<$\texttt{data}$>$} row data 
\STATE \textbf{output}: \texttt{features$_{\texttt{cat}}$}, \texttt{features$_{\texttt{num}}$}

\vspace{0.4cm}

 \STATE \textbf{function} {pre-processing}{$($\texttt{data}$)$}
 \STATE \hspace{0.7cm} \textit{remove duplication data}
 \STATE \hspace{0.7cm} \textit{rebild (or drop) missing (or incomplete) values}
 \STATE \hspace{0.7cm} \textit{remove (redundant) features with zero (and low) variance}
 \STATE \textbf{return} {\texttt{data}}
 
 \vspace{0.4cm}
 \STATE \textbf{function} {feature selection}{$($\texttt{data}$)$}
 \STATE \hspace{0.7cm} \textit{run proposed adjusted chi-squared-test (or adjusted mutual information)}
 \STATE \textbf{return} \texttt{features$_{\texttt{cat}}$}, \texttt{features$_{\texttt{num}}$}

 \vspace{0.4cm}
 \STATE \textbf{function} {feature encoder}{$($\texttt{features$_{\texttt{cat}}$}$)$}
 \STATE \hspace{0.7cm} \textbf{for} each feature $i$ \textbf{in}  \texttt{features$_{\texttt{cat}}$}
 \STATE \hspace{0.7cm}  \hspace{0.7cm} \texttt{d$\_$value} $\Leftarrow$ number of distinct values for feature $i$
 \STATE \hspace{0.7cm} \hspace{0.7cm}  \textbf{if}  \texttt{d$\_$value}$>$\texttt{threshold}
 \STATE \hspace{0.7cm} \hspace{0.7cm} \hspace{0.7cm}  \textit{do string indexer}
  \STATE \hspace{0.7cm} \hspace{0.7cm} \hspace{0.7cm}  \textit{do feature hasher}
 \STATE \hspace{0.7cm} \hspace{0.7cm}  \textbf{else} ($<$\texttt{threshold}) 
 \STATE \hspace{0.7cm} \hspace{0.7cm} \hspace{0.7cm}  \textit{do one-hot-encoder}
 \STATE \hspace{0.7cm} \hspace{0.7cm}  \textbf{end if} 
 \STATE \hspace{0.7cm} \textbf{end for}
 \STATE \textbf{return} encoded \texttt{features$_{\texttt{cat}}$}
 
 \vspace{0.4cm}
 \STATE \textbf{function} {feature scaler}{$($\texttt{features$_{\texttt{num}}$}$)$}
 \STATE \hspace{0.7cm} \textit{do normalization}
 \STATE \textbf{return} normalized \texttt{features$_{\texttt{num}}$}

 \end{algorithmic}
\end{algorithm}


\subsection{Data cleaning and pre-processing}
While unreliable data has a highly destructive  effect on the performance, data  cleaning is the process of detecting and refining (or deleting) corrupt, outlier, anomaly or inaccurate data from a dataset.
It refers to identifying defective, inaccurate, erroneous, inconsistent or irrelevant parts of the data, prior to replacing, modifying, removing or even reclaiming the dirty or coarse data, as well as removing any duplicate data.
In simple words, the data cleaning converts  data from an original raw form into a more convenient format.

Typically, in the case of digital marketing, when we receive the data, there are lots of duplication values, because of lack of centralized and accurate data gathering, recording or perfect online report generator tools. 
Of course, knowing the source of duplication can help a lot in the cleaning process.
However, in the more autonomous way, for data cleaning, even we can rely on the historical data. 
%
Another stage in data cleaning is rebuilding missing or incomplete data, where there are different solutions depending on the kind of problem such as time series analysis, machine learning, etc, and it is very difficult to provide a general solution.
But, before doing the data cleaning task, we have to figure out the reason why data goes missing, whereas the missing values happen in different manners, such as at random or not at random.
Missing at random means that the data trends to be missing is not relevant to the missing data, but it is related to some of the observed data.
Additionally, some  missing values have nothing to accomplish with their hypothetical values or with the values of other features (i.e. variables). 
In the other hand, the missing data could be not at random.
For instance, people with high salaries generally do not want to reveal their incomes, or the females generally do not want to reveal their ages. Here the missing value in 'age' feature is impacted by the 'gender' feature. So, we have to be really careful before removing any missing data.
Finally, when we figure out why the data is missing, we can decide to abandon missing values, or to fill.

As a summary,  the most important key benefits of data cleaning in digital marketing are: 
\begin{itemize}
    \item[--] Accurate view of customers, users and audiences;
    The customers and online users are the exclusive sources of any analysis and mining task and are central to all business decision making. 
    However, they are always changing. Their natures, behaviours, likes and dislikes, their habits, as well as their expectations are in a constant stage of change.
    Hence,  we need to remain on top of those fluctuations in order to make smart decisions. 
    \item[--] Improve data integration; Integrating data is vital to gain a complete view of the customers, users and audiences, and when data is dirty and laden with duplicates and errors, data integration process will be difficult. 
    In typical, multiple channels and multiple departments often collect duplicate data for the same user. With data cleaning, we can omit the useless and duplicate data and integrate it more effectively.
    \item[--] Increases revenue and productivity
\end{itemize}

\subsection{Feature selection}
In feature selection, we rely on the filter methods, and we try to fit two proposed adjusted statistical measures (i.e. mutual information, chi square test) to the observed data, then we select the features with the highest statistics. 
Suppose we have a label variable (i.e. the event label) and some feature variables that describe the data. 
We calculate the statistical test between every feature variable and the label variable and observe the existence of a relationship between the variable and the label. If the label variable is independent of the feature variable, we can discard that feature variable.
In the following, we present the proposed statistical measures in detail.

\subsubsection{Adjusted chi-squared test}
A very popular feature selection method is chi-squared test ($\chi ^2$-test).
In statistics, the chi-squared test is applied to test the independence of two events (i.e. features), where two events X and Y are defined to be independent, if $P(XY) = P(X)P(Y)$ or, equivalently, $P(X\vert Y)=P(X)$ and $P(Y\vert X)=P(Y)$. 
The formula for the $\chi^2$ is defined as:
\begin{equation}
\chi^2_{df} = \displaystyle \sum \frac{(O_i - E_i)^2}{E_i}
\end{equation}
where the subscript $df$ is the degree of freedom, $O$ is the observed value and $E$ the expected value.
The degrees of freedom ($df$) is equal to:
\begin{equation}
df = (r - 1) * (c - 1)
\end{equation}
where $r$ is the number of levels for one categorical feature, and $c$ is the number of levels for the other categorical feature.
After taking the following chi-square statistic, we need to find p-value in the chi-squared table, and decide whether to accept or reject the null hypothesis (H$_0$).  The  p-value is the probability of observing a sample statistic as extreme as the test statistic, and the null hypothesis is the case that two categorical features are independent.
Generally, small p-values reject the null hypothesis, and very large p-values means that the null hypothesis should not be rejected.
As result, the chi-squared test gives a p-value, which tells if the test results are significant or not.
In order to perform a chi-squared hypothesis test and get the p-value, one need a) the degree of freedom and b) the level of significance ($\alpha$), while the default value is 0.05 (5$\%$).

Like all non-parametric data, the chi-squared test is robust with respect to the distribution of the data \cite{McHugh2013}.
However, it has difficulty of interpretation when there are large numbers of categories in the features, and tendency to produce relatively very low p-values, even for insignificant features.
Furthermore, chi-squared test is sensitive to sample size, which is why several approaches to handle large data have been developed \cite{Bergh2015}.
When the cardinality is low, it would be a little difficult to get a null hypothesis rejection, whereas a higher cardinality will be more intended to result a rejection.

In feature selection, we usually calculate the p-values of each feature, then choose those ones which are smaller than the 'preset' threshold. Normally, we use $\alpha=0.95$, which stands for a threshold of 5\% significance. For those features within this threshold, smaller p-value stands for better feature. However as mentioned before, higher cardinality will always cause a lower p-value. This means that the features with higher cardinality, $e.g.$ \textit{user identification}, or \textit{site URLs}, are always having lower p-values, and in turn, to be a better feature, which may not always be true.

In order to find a more reliable measure other than simply using p-value from chi-squared test, we proposed a new measure by adding a regularization term on the p-values ($pv$) of features, called 'adjusted p-value' ($p_{\text{adj}}$). The new proposed statistical measure, $p_{\text{adj}}$, is defined as:

\begin{equation}
p_{\text{adj}} = \frac{\chi^2_{1-pv, df}  \, - \, \chi^2_{1-\alpha, df}}{\chi^2_{1-\alpha, df}}
\end{equation}
where $\alpha$ is the level of significance, and $df$ is the degrees of freedom.
By using this quantity, we are penalizing on the features with higher cardinality. Simply to say, we are trying to see how further by percentage the critical value corresponding to $pv$, the $\chi^2_{1-pv, df}$, is crossing the critical value $\chi^2_{1-\alpha, df}$, corresponding to a given significance level $1-\alpha$. Note that, the $\chi^2_{1-\alpha, df}$ will be very big for high cardinality features due to higher degree of freedom, and it is regarded as a penalization term.

The penalization could also be softer, if we take the logarithm of the critical value $\chi^2_{1-\alpha, df}$. In the case, the adjusted p-value with soft penalization, $\hat{p}_{\text{adj}}$, can be formulated as:

\begin{equation}
\hat{p}_{\text{adj}} = \frac{\chi^2_{1-pv, df} \, - \, \chi^2_{1-\alpha, df}}{\log(\chi^2_{1-\alpha, df})}
\end{equation}
For the two proposed above measures, higher value stands for better feature.

\subsubsection{Adjusted mutual information}
Similar to the Chi-square test, the Mutual Information (MI) is a statistic quanitity which measures how much a categorical variable tells another (mutual dependence between the two variables). 
The mutual information has two main properties; a) it can measure any kind of relationship between random variables, including nonlinear relationships \cite{Cover2006}, and b) it is invariant under the transformations in the feature space that are invertible and differentiable \cite{kullback1997}.
Therefore, it has been addressed in various kinds of studies with respect to feature selection \cite{CANG2012691,Vergara2014}.  It is formulated as:
\begin{equation}
I({X};{Y}) = \displaystyle \sum_{x \in {X}, y \in {Y}} P(x,y) \, \log\Big( \frac{P(x,y)}{P(x)P(y)} \Big)
\end{equation}
where  $I({X};{Y})$ stands for the 'mutual information' between two discrete variables ${X}$ and  ${Y}$, $P(x,y)$ is the joint probability of ${X}$ and  ${Y}$, and $P(x)$ and $P(y)$ are the marginal probability distribution of X and Y, respectively. 
The MI measure is a non-negative value, and it is easy to deduce that if ${X}$ is completely statistically independent from ${Y}$, we will get $P(x,y) = P(x)P(y)$ $s.t.$ $x \in {X}, y \in {Y}$, which indicates a MI value of 0.
The MI is bigger than 0, if ${X}$ is not independent from ${Y}$.

In simple words, the mutual information measures how much knowing one of the variables reduces uncertainty about the other one. For example, if $X$ and $Y$ are independent, then knowing $X$ does not give any information about $Y$ and vice versa, so their mutual information is zero.

In the case of feature selection using the label column (Y), if MI is equal to 0, then ${X}$ is considered as a 'bad' single feature, while the bigger MI value suggests more information provided from the feature ${X}$, which should be remained in the classification (predictive) model. 
Furthermore, when it comes to optimal feature subset, one can maximize the mutual information between the subset of selected features \textbf{x}$_{\text{subset}}$ and the label variable $Y$, as:

\begin{equation}
\zeta 
= \displaystyle arg\max_{\text{subset}} \, I(\textbf{x}_{\text{subset}};Y)
\end{equation}
where $|\text{subset}|=k$, and $k$ is the number of features to select.
The quantity $\zeta$ is called 'joint mutual information', and its maximizing is an $NP$-hard optimization problem.

However, the mutual information is subject to a  Chi-square distribution, that means we can convert the mutual information to a p-value as a new quantity. This new adjusted measure, called MI$_{\text{adj}}$, will be more robust than the standard mutual information. Also, we can rule out those features that is not significant based on the calculated p-value. Normally:

\begin{equation}
2N*\text{MI} \sim \chi^2_{df}
\end{equation}
where the $N$ stands for the number of data samples, and similar as before, $df$ is degree of freedom. So simply to say, our proposed new filtering rule (MI$_{\text{adj}}$) is defined by:

\begin{equation}
2N*I({X};{Y}) - \chi_{0.95, df(X,Y)} > 0
\end{equation}

The bigger MI$_{\text{adj}}$, the better is.
Some features will be ruled out if their new adjusted measures are negative, which indicate the mutual information are not significant comparing to their degrees of freedom.

\subsection{Feature encoding}

In the next step, we need to format the data, which can be accepted by the training model.
In practice, we do the training and prediction tasks using Spark in Yarn, because we have nearly forty million records to analyse on a weekly basis.
Specifically, we use the StringIndexer, which encodes the features by the rank of their occurrence times, for high cardinality features above a predefined threshold 
, and one hot encoder for the features whose unique levels less then the predefined threshold.
We also hash the high cardinality features to ensure that we are formatting the data without loosing too much information. 
In a nutshell, in our ad event prediction case, there are some extremely high cardinality features like user ids, or page urls with millions levels on weekly basis. 
It's better to hash them (rather than one hot encode them) to keep most of the information without facing the risk of explosion of feature numbers at the meantime.

\subsection{Feature scaling}
Finally, we do feature scaling according to the max-min scaling method, as a last step in the feature engineering.

\section{Experimental study}
In this section, we first describe the dataset used to conduct our experiments, then specify the validation process, prior to present and discuss the results that we obtained.

\subsection{Data description}
In this section, to clarify our claim in ad event prediction, we used a large real-world dataset of a running marketing campaign. 
The dataset is a private activity report from MediaMath digital advertising platform, is very huge, and the entire dataset is stored on cloud storage (i.e. Amazon S3) of Amazon Web Services (AWS). It comprises over 40 millions of recorded ads data (on weekly basis), each one with more than 80 pieces of information, which can be categorized in two main group: a) attributes which can be considered as features, and b) event labels for machine learning and mining tasks.


\subsubsection{Attributes (features)}
The input features for machine learning algorithms, such as \textit{user id}, \textit{site url},  \textit{browser}, \textit{date}, \textit{time}, \textit{location}, \textit{advertiser}, \textit{publisher} and \textit{channel}. Notice, that it is highly probable that one generates new features from the existence ones. For instance,  from \textit{start-time} and \textit{stop-time}, we can produce the \textit{duration} feature, or from \textit{date}, we can generate the new feature \textit{day of week}, which will be more meaningful in the advertising campaign.   

\subsubsection{Measures (labels)}
Measures are the target variables data which acts as labels in machine learning and data mining algorithms, such as  \textit{impressions}, \textit{clicks}, \textit{conversions}, \textit{videos}, and \textit{spend}. Note that, the mentioned measures (i.e. labels) are needed for supervised learning algorithms, while in non-supervised algorithm one can ignore them.

\subsection{Validation process}
Here we compare the proposed ad event prediction algorithm with the  state-of-the-art and the well-used feature engineering based event prediction methods.
For our comparisons, we rely on the Accuracy (ACC), recall or True Positive Rate (TPR),
 precision or Positive Predictive Value (PPV), 
  F-measure (F$_1$-SCORE), which is a harmonic mean of precision and recall, 
as well as the area under 
Precision-Recall curve (AUC-PR),
 which are commonly used in the literature, to evaluate each method.

\setlength{\tabcolsep}{10.2pt}
\begin{table}[!ht]
\linespread{1.2}
\centering
\caption{Confusion matrix}
{
\begin{tabular}{|cc|cc|}
\hline
&         & \multicolumn{2}{c|}{Predicted class } \\ \cline{3-4} 
&         &  Positive class      &  Negative class       \\ \hline
\multicolumn{1}{|c|}{\multirow{2}{*}{\rotatebox[origin=c]{0}{Actual class}}} & Positive class & TP          &  FN            \\
\multicolumn{1}{|c|}{}                        & Negative class & FP          &  TN           \\ \hline
\end{tabular}
\label{tab:confusionmatrix}
}
\end{table}
\linespread{1}

Table \ref{tab:confusionmatrix} presents the classical confusion matrix in terms of True Positive (TP), False Positive (FP), True Negative (TN) and False Negative (FN) values, which are used in the  performance metrics (P: event, N: normal).
Lastly, the above-mentioned comparison measures  are defined as:
\begin{equation}
{\mbox{ACC}} = \frac{TP+TN}{TP+FN+FP+TN}, \, \, {\mbox{TPR}} = \frac{TP}{TP+FN}, \, \, {\mbox{PPV}} = \frac{TP}{TP+FP}
\end{equation}



and,
\begin{equation}
{\mbox{F$_1$-SCORE}} = \frac{2TP}{2TP+FP+FN}
\end{equation}

The accuracy measure lies in [0, 100] in percentage, and true positive rate (recall), positive predictive value (precision),  and  F-measure lie within a range of  [0, 1]. 
The higher index, the better the agreement is. 
In the other side, 
precision-recall is a useful measure of success of prediction when the classes are very imbalanced, and the precision-recall curve shows the trade-off between precision and recall for different threshold. 
A high area under the curve represents both high recall and high precision.
High scores  illustrate that the predictor (or classifier) is returning accurate results (high precision), as well as returning a majority of all positive results (high recall).
The Precision-Recall (PR) summarizes such a curve as the weighted mean of precisions achieved at each threshold, with the increase in recall from the previous threshold used as the weight:
\begin{equation}
{\mbox{PR}} = \displaystyle \sum_n (R_n - R_{n-1}) P_n 
\end{equation}
where $P_n$ and $R_n$ are the precision and recall at the $n$th threshold.

For all the methods, the parameters as well as the training  and testing sets are formed by k-fold cross validation in the ratio of 80$\%$ and 20$\%$ of the data, respectively.  
For instance, for the random forest two parameters are tuned: number of trees and minimum sample leaf size.
Finally, the results reported hereinafter are averaged after 10 repetitions of the corresponding algorithm.

\subsection{Experimental results}
In the context of event prediction, the accuracy (ACC), 
recall (TPR), precision (PPV), F-measure (F$_1$-SCORE), as well as the area under 
PR curve (AUC-PR), for the various tested approaches, are reported in the Table \ref{tab_res}.
Many papers in the literature have shown
that, as far as heterogeneous multivariate data are concerned,
random forest are among the most efficient methods to be
considered \cite{Fernandez-Delgado:2014:WNH:2627435.2697065,Wainberg:2016:RFT:2946645.3007063}. Hence, we build our proposed event prediction algorithm on the basis of a Random Forest (RF) classifier.
To facilitate the big data analysis task, we do the data pre-processing, training, and ad event prediction by running spark jobs on Yarn. 
Note that, the results in bold correspond to the best assessment values.

\setlength{\tabcolsep}{8.87pt}
\begin{table}[!ht]
\caption{Comparison of performances based on the different feature engineering methods}
\label{tab_res}
\begin{tabular}{|l|c|c|c|c|c|}
\hline
\textbf{}                                            & {ACC} & {TPR} & {PPV} & {F$_1$-SCORE} & {AUC-PR} \\ \hline
{RF}                               &     99.61        &     0.050        &  0.013           &           0.022       &        0.005        \\ \hline
{RF + Feature Eng. ($\chi^2$)}          &   99.92           &   0.000           &     0.000         &     0.000                   &    0.005             \\ \hline
{RF + Feature Eng. ($\chi^2$-${p}_{\text{adj}}$)} &    \textbf{99.94}           &     \textbf{0.051}         &     \textbf{0.044}         &         \textbf{0.047}                   &    \textbf{0.010}             \\ \hline
{RF + Feature Eng. ($\chi^2$-$\hat{p}_{\text{adj}}$)} &    99.92           &     0.000         &         0.000     &         0.000                 &    0.006             \\ \hline
{RF + Feature Eng. (MI)}           &     99.92           &    0.000          &    0.000          &        0.000                        &    0.006          \\ \hline
{RF + Feature Eng. (MI$_{\text{adj}}$)}  &   99.91         &      0.008        &      0.023        &          0.012               &    0.006             \\ \hline

\end{tabular}
\end{table}

Table \ref{tab_res} shows the comparison of performances of a simple classifier versus the case of doing a feature engineering before running the classifier, on a real-world dataset. 
Notice that, here, in the feature selection process,  we consider only top 20 features. Of course, one can simply find the best number of features using a $k$-fold cross validation technique. 

As has been pointed out in Table \ref{tab_res}, while in feature selection using the standard statistical approach (i.e. $\chi^2$ and MI), the random forest classifier can not provide any good results for recall (TPR), precision (PPV) and F$_1$-SCORE, using the proposed statistical metrics (i.e. $\chi^2$-${p}_{\text{adj}}$ and MI$_{\text{adj}}$), we generally outperform the results.
Also, it is plain to see that the RF classifier with considering feature selection based on the proposed $\chi^2$-${p}_{\text{adj}}$ has the best results, and outperforms significantly the precision, recall, F-measure as well as the area under precision-recall curve.
However, for this case, the soft penalized version of $\chi^2$-${p}_{\text{adj}}$ (i.e. $\chi^2$-$\hat{p}_{\text{adj}}$) does not provide very good result, but still it is better than the standard $\chi^2$ and is comparable with the standard mutual information metric.
Furthermore, as demonstrated, using the proposed adjusted version of mutual information in feature selection process, provides better results rather than using the standard mutual information measure.

To verify our claim and consolidate the comparative results, we use a Wilcoxon signed rank test, which is a non-parametric statistical hypothesis test to effectively determine whether our proposed adjusted statistic measures are significantly outperform the classifier (using the alternative quantities) or not. 
Tables \ref{tab_sign} presents the two-sided p-value for the hypothesis test,
while the results in bold indicate the significantly different ones. 
The p-value is the probability of observing a test statistic more extreme than the observed value under the null hypothesis. Notice that, the null hypothesis is strongly rejected when the p-values are lower than a pre-defined criterion, almost always set to 0.05.
It means that the differences between the two tested classifiers are significant and the uniform hypothesis is accepted as p-values are greater than 0.05. 
Based on the p-values, we can justify that using the proposed adjusted measures in feature selection, the classifier leads to significantly better results than the others. 
Note that the difference between the pairs of classifiers results follows a symmetric distribution around zero and to be more precise, the reported p-values are
computed from all the individual results after some repetitions of the corresponding algorithm.

\setlength{\tabcolsep}{9.6pt}
\begin{table}[!hb]
\caption{P-values: Wilcoxon test}
\label{tab_sign}
\begin{tabular}{|l|c|c|c|c|c|}
\hline
\multirow{2}{*}{} & \multicolumn{5}{c|}{RF + Feature Eng.} \\ \cline{2-6} 
                  &    ($\chi^2$)    &   ($\chi^2$-${p}_{\text{adj}}$)    &   ($\chi^2$-$\hat{p}_{\text{adj}}$)    &    (MI)   &  (MI$_{\text{adj}}$)     \\ \hline
RF              &       \textbf{0.034}         &    \textbf{0.004}             &     0.073            &      0.073           &   0.798              \\ \hline
RF + Feature Eng. ($\chi^2$) &               &   \textbf{0.011}              &    0.157             &     0.157            &      \textbf{0.011}           \\ \hline
RF + Feature Eng. ($\chi^2$-${p}_{\text{adj}}$) &                 &                &   \textbf{0.011}              &        \textbf{0.011}         &         \textbf{0.011}        \\ \hline
RF + Feature Eng. ($\chi^2$-$\hat{p}_{\text{adj}}$)          &      &                 &             &  1.000               &       \textbf{0.026}          \\ \hline
RF + Feature Eng. (MI) &      &                  &               &                &       \textbf{0.026}         \\ \hline
\end{tabular}
\end{table}

As is evident from Table \ref{tab_sign}, with regard to the p-values of Wilcoxon test, the RF classifier with considering feature selection method using the proposed $\chi^2$-${p}_{\text{adj}}$ measure, brings a significant improvement compared to the other methods.
In addition, the proposed MI$_{\text{adj}}$ is almost performing significantly different than the other approaches.

To become closely acquainted with selected features, Table \ref{tab_features} shows the top 20 selected features using different feature selection models (i.e. statistical test) which we consider in our experimental study. 

\setlength{\tabcolsep}{.1pt}
\begin{table}[!ht]
\centering
\small
{
\caption{Feature selection: top-$k$ extracted features using different statistical test (rank sorted)}
\label{tab_features}
\scalebox{0.97}{
    \begin{tabular}{|c|c|c|c|c|}
      \hline
    $\chi^2$   &   $\chi^2$-${p}_{\text{adj}}$   &   $\chi^2$-$\hat{p}_{\text{adj}}$    &    MI  & MI$_{\text{adj}}$    \\
      \hline
       \texttt{os\_id} & \texttt{id\_vintage} & \texttt{mm\_uuid} &   \texttt{mm\_uuid} & \texttt{publisher\_id} \\
       \texttt{category\_id} & \texttt{os\_id} & \texttt{contextual\_data} & \texttt{contextual\_data} & \texttt{site\_id} \\
       \texttt{contextual\_data} & \texttt{fold\_position} & \texttt{page\_url} & \texttt{page\_url} & \texttt{site\_url} \\
       \texttt{week\_hour} & \texttt{advertiser\_id} & \texttt{user\_agent} & \texttt{site\_id} & \texttt{main\_page} \\
       \texttt{concept\_id} & \texttt{conn\_speed\_id} & \texttt{strategy\_id} & \texttt{strategy\_id} & \texttt{supply\_source\_id} \\
       \texttt{overlapped\_b\_pxl} & \texttt{watermark} & \texttt{creative\_id} & \texttt{site\_url} & \texttt{exchange\_id} \\
       \texttt{week\_hour\_part} & \texttt{browser\_name} & \texttt{site\_id} & \texttt{publisher\_id} & \texttt{category\_id} \\
       \texttt{site\_id} & \texttt{browser\_language} & \texttt{zip\_code\_id} & \texttt{creative\_id} & \texttt{creative\_id} \\
       \texttt{norm\_rr} & \texttt{week\_hour} & \texttt{ip\_address} & \texttt{user\_agent} & \texttt{channel\_type} \\
       \texttt{page\_url} & \texttt{width} & \texttt{site\_url} & \texttt{main\_page} & \texttt{concept\_id} \\
       \texttt{strategy\_id} & \texttt{cross\_device\_flag} & \texttt{overlapped\_b\_pxl} & \texttt{zip\_code\_id} & \texttt{campaign\_id} \\
       \texttt{app\_id} & \texttt{browser\_version} & \texttt{aib\_recencies} & \texttt{concept\_id} & \texttt{browser\_name} \\
       \texttt{creative\_id} & \texttt{week\_hour\_part} & \texttt{main\_page} & \texttt{campaign\_id} & \texttt{browser\_version} \\
       \texttt{browser\_version} & \texttt{device\_id} & \texttt{city\_code\_id} & \texttt{city\_code\_id} & \texttt{model\_name} \\
       \texttt{zip\_code\_id} & \texttt{app\_id} & \texttt{concept\_id} & \texttt{supply\_source\_id} & \texttt{homebiz\_type\_id} \\
       \texttt{form\_factor} & \texttt{city} & \texttt{campaign\_id} & \texttt{exchange\_id} & \texttt{os\_id} \\
       \texttt{supply\_source\_id} & \texttt{week\_part\_hour\_part} & \texttt{channel\_type} & \texttt{category\_id} & \texttt{os\_name} \\
       \texttt{prebid\_viewability} & \texttt{category\_id} & \texttt{homebiz\_type\_id} & \texttt{model\_name} & \texttt{brand\_name} \\
       \texttt{brand\_name} & \texttt{norm\_rr} & \texttt{os\_id} & \texttt{norm\_rr} & \texttt{form\_factor} \\
       \texttt{exchange\_id} & \texttt{dma\_id} & \texttt{form\_factor} & \texttt{channel\_type} & \texttt{norm\_rr} \\
      \hline
    \end{tabular}
}
}
\end{table}

Note that, in the term of time consumption, all the proposed adjusted statistical test (i.e. $\chi^2$-${p}_{\text{adj}}$, $\chi^2$-$\hat{p}_{\text{adj}}$,  MI$_{\text{adj}}$  
), have more or less the same time complexity with the standard ones (i.e. $\chi^2$, MI), without any significant difference. 

\begin{figure}[!ht]
\centerline
{
\includegraphics[width=7.57cm]{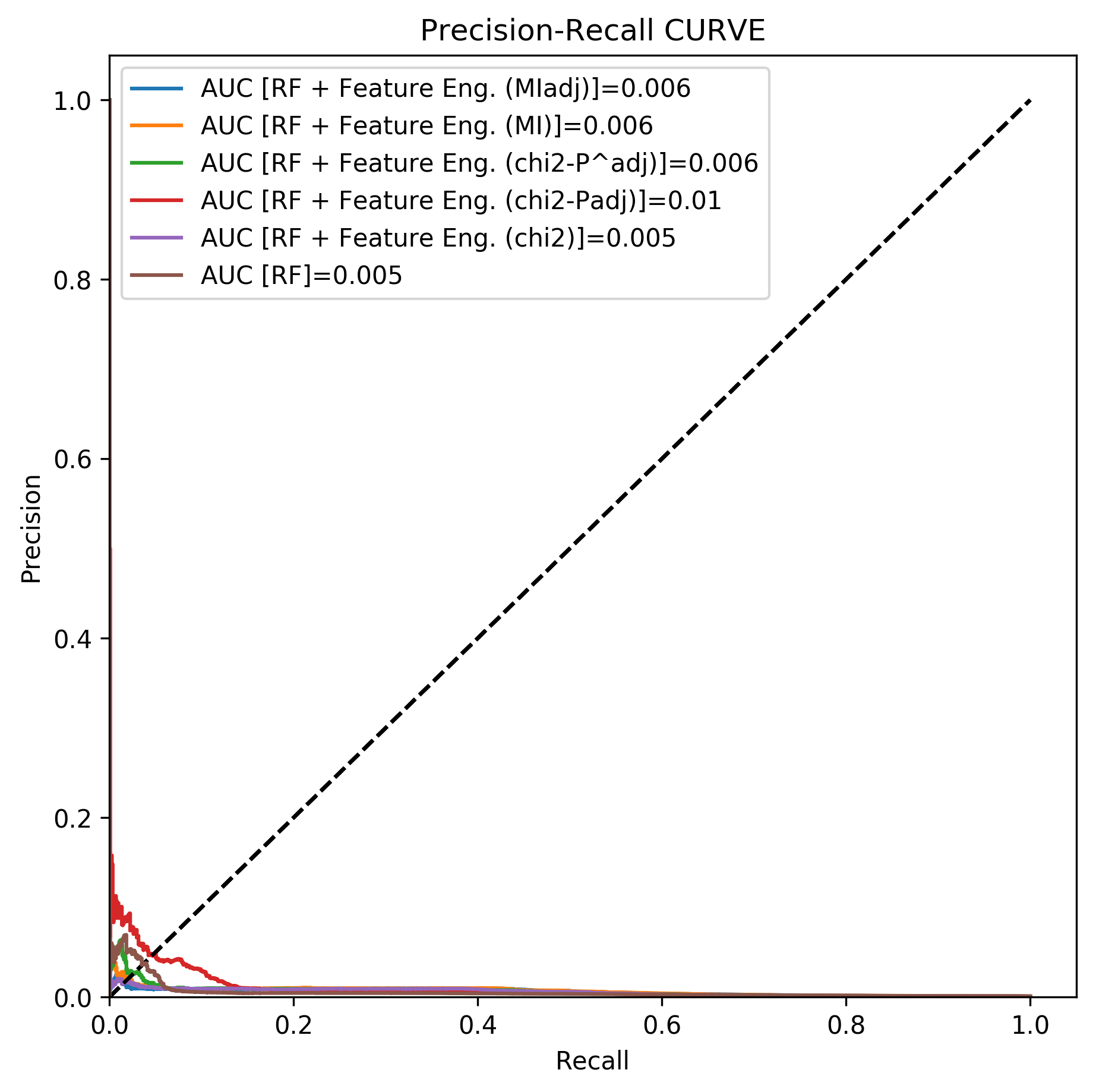} \\
\includegraphics[width=7.57cm]{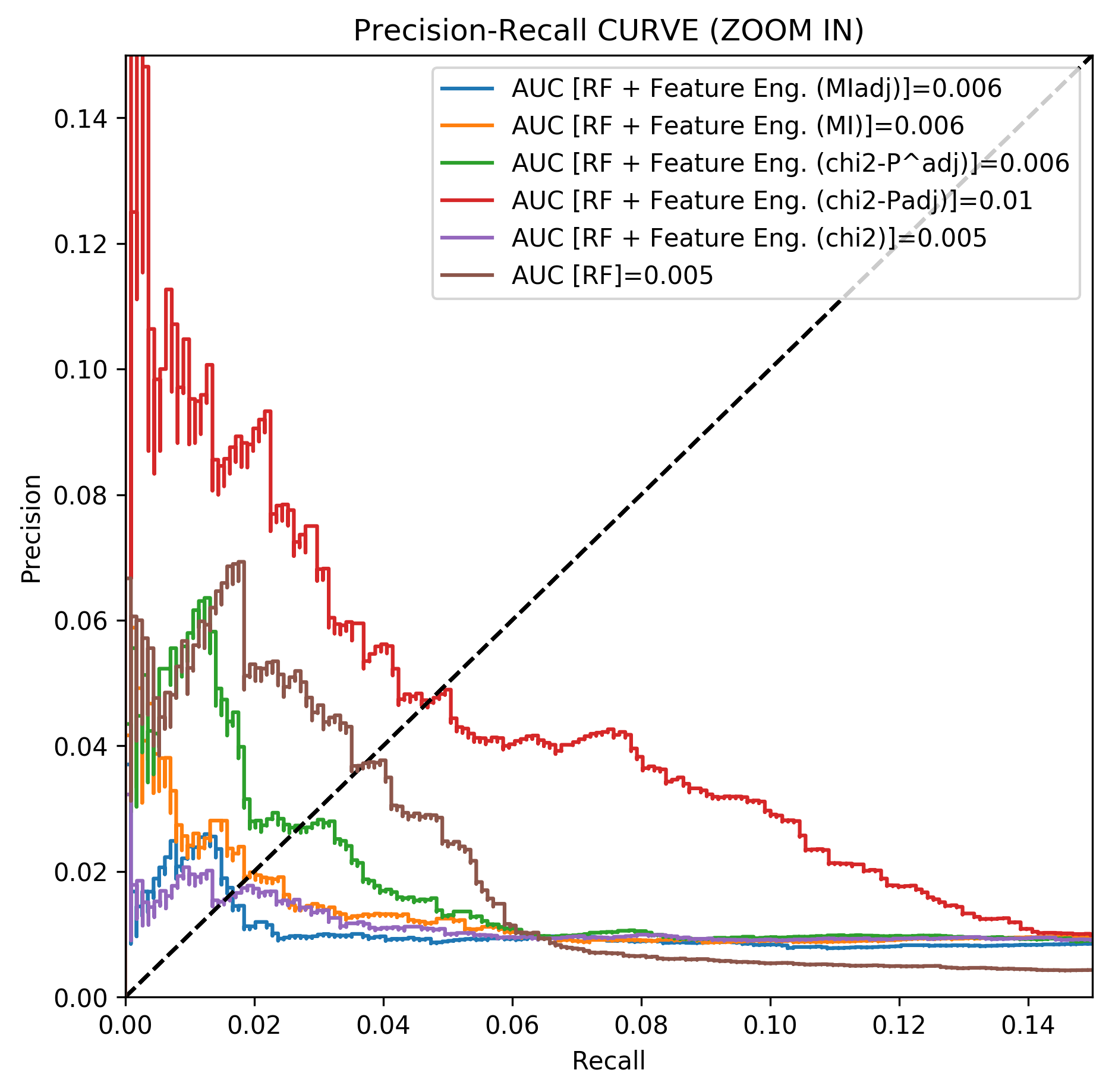} 
}
\caption{Comparison of AUC-PR curve based of different feature engineering methods}
\label{fig:comp_aup}
\end{figure}

Lastly, to have a closer look at the ability of the proposed statistical quantities, the Figure \ref{fig:comp_aup} shows the comparison of the area under PR curve for different feature engineering methods on the basis of a random forest classifier.
Note that, we consider the area under PR curve, since it is more reliable rather than ROC curve, because of the unbalanced nature of data.  
The higher values are the better performance.

\section{Conclusion}
\label{sec:conclude}
This research work introduces an enhanced ad event prediction framework which has been applied on big data. In this framework, we propose two statistical approaches which can be used for feature selection: i) the adjusted Chi-squared test and ii) the adjusted mutual information. Then, by ranking the statistical measures we select the best features. 
Also, in feature encoding before training the model, we used a practical while reliable pipeline to encode very large (categorical) data. 
The efficiency of the proposed  event prediction framework is analyzed on a large real-world dataset from a running campaign.
The results illustrate the benefits of the proposed adjusted Chi-squared test (and the adjusted mutual information), which outperforms the others with respect to different metrics, i.e. accuracy, precision, recall, F-measure and the area under precision-recall curve. 
Lastly, a Wilcoxon signed-rank test is used to determine that the proposed approach is significantly better than the other methods described in the paper.
While, in this research work, we focus on the single features, the idea of combined features can be a proper proposition to gain better result.
Hence, investigate the combination of some features to generate more useful features, to further increase the prediction performance of the imbalanced case, which is typical in the context of digital advertising,  can be an interesting suggestion for future works.

\bibliographystyle{ieeetr}

{\small
\bibliography{soheily_wu}}

\begin{thebibliography}{10}

\bibitem{Cheng:2010:PCP:1718487.1718531}
H.~Cheng and E.~Cant\'{u}-Paz, ``Personalized click prediction in sponsored
  search,'' in {\em Proceedings of the Third ACM International Conference on
  Web Search and Data Mining}, WSDM '10, (New York, NY, USA), pp.~351--360,
  ACM, 2010.

\bibitem{Zhang:2014:SCP:2893873.2894086}
Y.~Zhang, H.~Dai, C.~Xu, J.~Feng, T.~Wang, J.~Bian, B.~Wang, and T.-Y. Liu,
  ``Sequential click prediction for sponsored search with recurrent neural
  networks,'' in {\em Proceedings of the Twenty-Eighth AAAI Conference on
  Artificial Intelligence}, pp.~1369--1375, AAAI Press, 2014.

\bibitem{Chapelle:2014:SSR:2699158.2532128}
O.~Chapelle, E.~Manavoglu, and R.~Rosales, ``Simple and scalable response
  prediction for display advertising,'' {\em ACM Trans. Intell. Syst.
  Technol.}, vol.~5, pp.~61:1--61:34, Dec. 2014.

\bibitem{Borisov:2016:NCM:2872427.2883033}
A.~Borisov, I.~Markov, M.~de~Rijke, and P.~Serdyukov, ``A neural click model
  for web search,'' in {\em Proceedings of the 25th International Conference on
  World Wide Web}, pp.~531--541, 2016.

\bibitem{Cheng:2016:WDL:2988450.2988454}
H.-T. Cheng, L.~Koc, J.~Harmsen, T.~Shaked, T.~Chandra, H.~Aradhye,
  G.~Anderson, G.~Corrado, W.~Chai, M.~Ispir, R.~Anil, Z.~Haque, L.~Hong,
  V.~Jain, X.~Liu, and H.~Shah, ``Wide \& deep learning for recommender
  systems,'' in {\em Proceedings of the 1st Workshop on Deep Learning for
  Recommender Systems}, DLRS 2016, (New York, NY, USA), pp.~7--10, ACM, 2016.

\bibitem{Li:2015:CPA:2783258.2788582}
C.~Li, Y.~Lu, Q.~Mei, D.~Wang, and S.~Pandey, ``Click-through prediction for
  advertising in twitter timeline,'' in {\em Proceedings of the 21th ACM SIGKDD
  International Conference on Knowledge Discovery and Data Mining}, KDD '15,
  pp.~1959--1968, ACM, 2015.

\bibitem{Chen:2016:DCP:2964284.2964325}
J.~Chen, B.~Sun, H.~Li, H.~Lu, and X.-S. Hua, ``Deep ctr prediction in display
  advertising,'' in {\em Proceedings of the 24th ACM International Conference
  on Multimedia}, MM '16, (New York, NY, USA), pp.~811--820, ACM, 2016.

\bibitem{Richardson:2007:PCE:1242572.1242643}
M.~Richardson, E.~Dominowska, and R.~Ragno, ``Predicting clicks: Estimating the
  click-through rate for new ads,'' in {\em Proceedings of the 16th
  International Conference on World Wide Web}, WWW '07, (New York, NY, USA),
  pp.~521--530, ACM, 2007.

\bibitem{citeulike:4375063}
D.~Agarwal, B.~C. Chen, and P.~Elango, ``Spatio-temporal models for estimating
  click-through rate,'' in {\em WWW '09: Proceedings of the 18th international
  conference on World wide web}, (New York, NY, USA), pp.~21--30, ACM, 2009.

\bibitem{Graepel:2010:WBC:3104322.3104326}
T.~Graepel, J.~Q.~n. Candela, T.~Borchert, and R.~Herbrich, ``Web-scale
  bayesian click-through rate prediction for sponsored search advertising in
  microsoft's bing search engine,'' in {\em Proceedings of the 27th
  International Conference on International Conference on Machine Learning},
  ICML'10, (USA), pp.~13--20, Omnipress, 2010.

\bibitem{Chapelle:2014:MDF:2623330.2623634}
O.~Chapelle, ``Modeling delayed feedback in display advertising,'' in {\em
  Proceedings of the 20th ACM SIGKDD International Conference on Knowledge
  Discovery and Data Mining}, KDD '14, (New York, NY, USA), pp.~1097--1105,
  ACM, 2014.

\bibitem{Effendi2017ClickTR}
M.~J. Effendi and S.~A. Ali, ``Click through rate prediction for contextual
  advertisment using linear regression,'' {\em CoRR}, vol.~abs/1701.08744,
  2017.

\bibitem{41159}
H.~B. McMahan, G.~Holt, D.~Sculley, M.~Young, D.~Ebner, J.~Grady, L.~Nie,
  T.~Phillips, E.~Davydov, D.~Golovin, S.~Chikkerur, D.~Liu, M.~Wattenberg,
  A.~M. Hrafnkelsson, T.~Boulos, and J.~Kubica, ``Ad click prediction: a view
  from the trenches,'' in {\em Proceedings of the 19th ACM SIGKDD International
  Conference on Knowledge Discovery and Data Mining (KDD)}, 2013.

\bibitem{Friedman:2002:SGB:635939.635941}
J.~H. Friedman, ``Stochastic gradient boosting,'' {\em Comput. Stat. Data
  Anal.}, vol.~38, pp.~367--378, Feb. 2002.

\bibitem{Trofimov:2012:UBT:2351356.2351358}
I.~Trofimov, A.~Kornetova, and V.~Topinskiy, ``Using boosted trees for
  click-through rate prediction for sponsored search,'' in {\em Proceedings of
  the Sixth International Workshop on Data Mining for Online Advertising and
  Internet Economy}, ADKDD '12, (New York, NY, USA), pp.~2:1--2:6, ACM, 2012.

\bibitem{burges2010ranknet}
C.~J.~C. Burges, ``From {RankNet} to {LambdaRank} to {LambdaMART}: An
  overview,'' tech. rep., Microsoft Research, 2010.

\bibitem{Dave:2010:LCR:1835449.1835671}
K.~S. Dave and V.~Varma, ``Learning the click-through rate for rare/new ads
  from similar ads,'' in {\em Proceedings of the 33rd International ACM SIGIR
  Conference on Research and Development in Information Retrieval}, SIGIR '10,
  (New York, NY, USA), pp.~897--898, ACM, 2010.

\bibitem{Domingos1997}
P.~Domingos and M.~Pazzani, ``On the optimality of the simple bayesian
  classifier under zero-one loss,'' {\em Machine Learning}, vol.~29, no.~2,
  pp.~103--130, 1997.

\bibitem{journals/jcit/Entezari-MalekiRM09}
R.~Entezari-Maleki, A.~Rezaei, and B.~Minaei-Bidgoli, ``Comparison of
  classification methods based on the type of attributes and sample size.,''
  {\em JCIT}, vol.~4, no.~3, pp.~94--102, 2009.

\bibitem{journals/bmcbi/KukrejaJS12}
M.~Kukreja, S.~A. Johnston, and P.~Stafford, ``Comparative study of
  classification algorithms for immunosignaturing data.,'' {\em BMC
  Bioinformatics}, vol.~13, p.~139, 2012.

\bibitem{Lorena20115268}
A.~C. Lorena, L.~F. Jacintho, M.~F. Siqueira, R.~D. Giovanni, L.~G. Lohmann,
  A.~C. de~Carvalho, and M.~Yamamoto, ``Comparing machine learning classifiers
  in potential distribution modelling,'' {\em Expert Systems with
  Applications}, vol.~38, no.~5, pp.~5268 -- 5275, 2011.

\bibitem{Zhou2017DeepIN}
G.~Zhou, C.~Song, X.~Zhu, X.~Ma, Y.~Yan, X.~Dai, H.~Zhu, J.~Jin, H.~Li, and
  K.~Gai, ``Deep interest network for click-through rate prediction,'' {\em
  CoRR}, vol.~abs/1706.06978, 2017.

\bibitem{Liu:2015:CCP:2806416.2806603}
Q.~Liu, F.~Yu, S.~Wu, and L.~Wang, ``A convolutional click prediction model,''
  in {\em Proceedings of the 24th ACM International on Conference on
  Information and Knowledge Management}, CIKM '15, (New York, NY, USA),
  pp.~1743--1746, ACM, 2015.

\bibitem{Zhang2016DeepLO}
W.~Zhang, T.~Du, and J.~Wang, ``Deep learning over multi-field categorical
  data: A case study on user response prediction,'' in {\em ECIR}, 2016.

\bibitem{Breiman2001}
L.~Breiman, ``Random forests,'' {\em Machine Learning}, vol.~45, no.~1,
  pp.~5--32, 2001.

\bibitem{liaw02:rf_pack}
A.~Liaw and M.~Wiener, ``Classification and regression by random forest,'' {\em
  R News}, vol.~2, no.~3, pp.~18--22, 2002.

\bibitem{8367767}
S.~{Soheily-Khah}, P.~{Marteau}, and N.~{B\'echet}, ``Intrusion detection in
  network systems through hybrid supervised and unsupervised machine learning
  process: A case study on the iscx dataset,'' in {\em 2018 1st International
  Conference on Data Intelligence and Security (ICDIS)}, pp.~219--226, April
  2018.

\bibitem{Dembczynski2008PredictingA}
K.~Dembczynski, W.~Kotlowski, and D.~Weiss, ``Predicting ads ’ click-through
  rate with decision rules,'' in {\em WWW2008, Beijing, China}, 2008.

\bibitem{Trofimov2012}
I.~Trofimov, A.~Kornetova, and V.~Topinskiy, ``Using boosted trees for
  click-through rate prediction for sponsored search,'' in {\em Proceedings of
  the Sixth International Workshop on Data Mining for Online Advertising and
  Internet Economy}, ADKDD '12, (New York, NY, USA), pp.~2:1--2:6, ACM, 2012.

\bibitem{shi2016}
L.~Shi and B.~Li, ``Predict the click-through rate and average cost per click
  for keywords using machine learning methodologies,'' in {\em Proceedings of
  the International Conference on Industrial Engineering and Operations
  ManagementDetroit, Michigan, USA}, 2016.

\bibitem{Shan:2016:DCW:2939672.2939704}
Y.~Shan, T.~R. Hoens, J.~Jiao, H.~Wang, D.~Yu, and J.~Mao, ``Deep crossing:
  Web-scale modeling without manually crafted combinatorial features,'' in {\em
  Proceedings of the 22Nd ACM SIGKDD International Conference on Knowledge
  Discovery and Data Mining}, KDD '16, (New York, NY, USA), pp.~255--262, ACM,
  2016.

\bibitem{soheily2018ensemble}
S.~{Soheily-Khah} and Y.~{Wu}, ``Ensemble learning using frequent itemset
  mining for anomaly detection,'' in {\em International Conference on
  Artificial Intelligence, Soft Computing and Applications (AIAA 2018)}, 2018.

\bibitem{McHugh2013}
M.~L. McHugh, ``The chi-square test of independence,'' {\em Biochemia Medica},
  vol.~23, p.~143–149, 2013.

\bibitem{Bergh2015}
D.~Bergh, ``Sample size and chi-squared test of fit : A comparison between a
  random sample approach and a chi-square value adjustment method using swedish
  adolescent data.,'' {\em In Pacific Rim Objective Measurement Symposium
  (PROMS) 2014 Conference Proceedings}, p.~197–211, 2015.

\bibitem{Cover2006}
T.~M. Cover and J.~A. Thomas, {\em Elements of Information Theory 2nd Edition
  (Wiley Series in Telecommunications and Signal Processing)}.
\newblock Wiley-Interscience, July 2006.

\bibitem{kullback1997}
S.~Kullback, {\em {Information Theory And Statistics}}.
\newblock Dover Pubns, 1997.

\bibitem{CANG2012691}
S.~Cang and H.~Yu, ``Mutual information based input feature selection for
  classification problems,'' {\em Decision Support Systems}, vol.~54, no.~1,
  pp.~691 -- 698, 2012.

\bibitem{Vergara2014}
J.~R. Vergara and P.~A. Est{\'e}vez, ``A review of feature selection methods
  based on mutual information,'' {\em Neural Computing and Applications},
  vol.~24, pp.~175--186, Jan. 2014.

\bibitem{Fernandez-Delgado:2014:WNH:2627435.2697065}
M.~Fern\'{a}ndez-Delgado, E.~Cernadas, S.~Barro, and D.~Amorim, ``Do we need
  hundreds of classifiers to solve real world classification problems?,'' {\em
  J. Mach. Learn. Res.}, vol.~15, pp.~3133--3181, Jan. 2014.

\bibitem{Wainberg:2016:RFT:2946645.3007063}
M.~Wainberg, B.~Alipanahi, and B.~J. Frey, ``Are random forests truly the best
  classifiers?,'' {\em J. Mach. Learn. Res.}, vol.~17, pp.~3837--3841, Jan.
  2016.

\end{thebibliography}

\vspace{1cm}

\textbf{Authors}

  \begin{wrapfigure}{r}{25mm} 
  \vspace{-0.39cm}
    \includegraphics[width=0.9in,height=1.15in,clip]{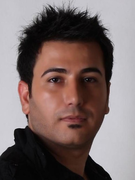}
  \end{wrapfigure}\par
  \textbf{Saeid SOHEILY KHAH}
  {\small graduated 
in software engineering,
and received master degree in artificial intelligence $\&$
robotics. 
He then received his second master degree in information analysis and management from Skarbek university, Warsaw. 
In 
2013, he joined to the LIG (Laboratoire d'Informatique de Grenoble) 
at Universit\'e Grenoble Alpes as a doctoral researcher. He successfully defended his dissertation and 
got his Ph.D in Oct 2016.
Instantly, he joined to the IRISA  at Université Bretagne Sud as a postdoctoral researcher. Lastly, in Oct 2017, he joined Skylads 
as a research scientist.
His research interests are machine learning, data mining, 
cyber security system, digital advertising and artificial intelligence.}\par

  \begin{wrapfigure}{r}{25mm} 
  \vspace{-0.39cm}
    \includegraphics[width=0.9in,height=1.15in,clip]{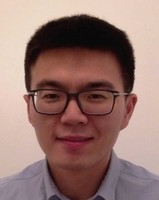}
  \end{wrapfigure}\par
  \textbf{Yiming WU} 
  {\small received his B.S.E.E. degree from the  Northwestern Polytechnical University, Xi'an in China. He received his Ph.D. degree in Electrical Engineering from University of Technology of Belfort-Montb\'eliard, Belfort, France, 2016. He joined Skylads as a data scientist in 2018, and his research has addressed topics on machine learning, artificial intelligence and digital advertising.} \par

\end{document}